\documentclass{article}
\usepackage{spconf,amsmath,amssymb,graphicx,booktabs,multirow,array}
\usepackage[table]{xcolor}
\usepackage{balance}
\usepackage{capt-of}
\usepackage[T1]{fontenc}
\usepackage{tikz}
\usepackage[hidelinks]{hyperref}
\definecolor{icRT}{HTML}{0072B2}\definecolor{icGP}{HTML}{56677A}\definecolor{icMD}{HTML}{D55E00}
\newcommand{\icBadge}[2]{\tikz[baseline=-0.6ex,x=1ex,y=1ex]{\fill[#1](0,0) circle (1.0);#2}}
\newcommand{\icTag}[2]{\tikz[baseline=(t.base)]{\node[inner sep=1.2pt,rounded corners=1pt,fill=#1!15,text=#1,font=\sffamily\scriptsize\bfseries](t){#2};}}
\newcommand{\icFast}{\icBadge{icRT}{\fill[white](0.15,0.65)--(-0.38,-0.05)--(-0.02,-0.05)--(-0.15,-0.65)--(0.38,0.1)--(0.02,0.1)--cycle;}\,\icTag{icRT}{RT}}
\newcommand{\icGeneral}{\icBadge{icGP}{\fill[white](-0.6,-0.4)--(-0.15,0.25)--(0.12,-0.08)--(0.3,0.1)--(0.6,-0.4)--cycle;\fill[white](0.28,0.42) circle (0.13);}\,\icTag{icGP}{Gen}}
\newcommand{\icMedical}{\icBadge{icMD}{\fill[white](-0.17,-0.58) rectangle (0.17,0.58);\fill[white](-0.58,-0.17) rectangle (0.58,0.17);}\,\icTag{icMD}{Med}}

\newcommand{\brow}{\rowcolor{blue!10}}


\title{Seeing the Small:\\Deformable Duality Mamba for Retinal Lesion Segmentation}
\name{Chang Yu, Siqing Li, Yuanchen Liu, Zhenyu Bu, Luming Yang$^{\ast}$, Alper Yilmaz}
\address{The Ohio State University, Columbus, OH, USA\\ $^{\ast}$Correspondence: yang.7670@osu.edu}

\begin{document}
\maketitle
\raggedbottom
\begin{abstract}
Diabetic retinopathy (DR) lesions vary widely in shape, and microaneurysms (MAs) span only a few pixels. We hypothesize that modeling the spatial organization of lesion regions, with global context and shape-adaptive scanning, improves the segmentation of such small and irregular lesions. To test this, we propose D$^2$Mamba, which couples non-causal state space duality with deformable Mamba scanning. We evaluate it with five-fold cross-validation on the DDR dataset, comparing against representative segmentation baselines under an identical training protocol, and conduct ablations on each component. D$^2$Mamba achieved the best overall performance, with 33.87\% mIoU, 2.99 percentage points above the strongest baseline. The largest gain appeared on MAs, the smallest and most easily missed lesions, where Dice improved by 10.06 percentage points. These results confirm our hypothesis that spatially adaptive state space modeling benefits tiny-lesion segmentation. Code and pretrained weights are available at {\color{blue}\url{http://github.com/PUREPANEL1/D2Mamba}}.
\end{abstract}

\section{Introduction}

\begin{figure}[!t]
\centering
\includegraphics[width=\columnwidth]{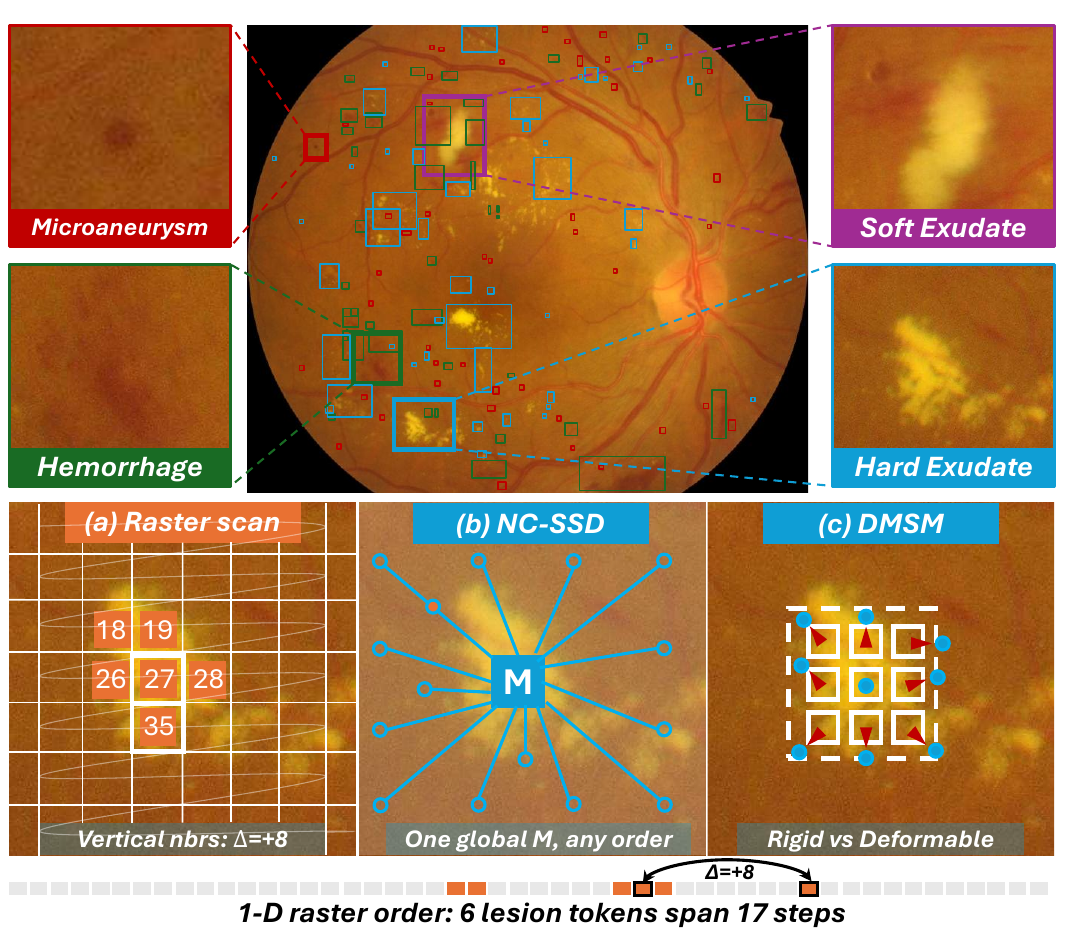}
\caption{\textbf{Lesion appearances and spatial modeling in case 49 of IDRiD dataset.} Colored boxes mark native-resolution crops of microaneurysms (MA), hemorrhages (HE), hard exudates (EX), and soft exudates (SE). \textbf{(a)} Raster scanning separates nearby lesion features in the sequence. \textbf{(b)} NC-SSD gathers global context without directional scanning. \textbf{(c)} DMSM adapts sampling to irregular lesion shapes.}
\label{fig:lesion_example}
\end{figure}

Diabetic retinopathy (DR) is a leading cause of preventable vision loss, affecting roughly one in five people with diabetes worldwide~\cite{cheung2010,teo2021}. Deep learning now supports referable-DR screening, including autonomous systems deployed in primary care~\cite{gulshan2016,ting2017,abramoff2018}, and retinal foundation models~\cite{retfound}; however, these systems output image-level grades rather than lesion maps. Lesion-level evidence consists of hard exudates (EX), microaneurysms (MA), soft exudates (SE), and hemorrhages (HE), annotated pixel-wise in public benchmarks~\cite{idrid2018,idrid2020,ddr}. Segmenting these lesions in fundus images is difficult: they differ in morphology, small lesions occupy a tiny fraction of an image, downsampling can discard their signal, and background pixels can dominate training (Fig.~\ref{fig:lesion_example}).

Networks with an encoder and decoder, such as U-Net and nnU-Net, are standard for biomedical segmentation~\cite{litjens2017,unet,nnunet}. General-purpose convolutional models trade detail against context in different ways: Fast-SCNN shares early downsampling~\cite{fastscnn}, CGNet combines local and surrounding features~\cite{cgnet}, BiSeNetV2 separates detail from semantics~\cite{bisenetv2}, and UPerNet fuses multi-level features~\cite{upernet}. For fundus lesions, L-Seg unifies multi-lesion segmentation~\cite{lseg} and HACDR-Net uses heterogeneous receptive fields for varied lesion geometry~\cite{hacdr}. Repeated downsampling can nevertheless weaken evidence from small lesions.

Non-local operations~\cite{dnl} and Transformers such as Swin-Unet and SegFormer~\cite{swinunet,segformer} enlarge the interaction range, but the additional context can include irrelevant retinal structure. Mamba~\cite{mamba} was designed for sequential data that evolve over time; visual variants such as VMamba~\cite{vmamba}, U-Mamba~\cite{umamba}, and MambaLiteUNet~\cite{mambaliteunet} instead apply its state transitions over spatial positions. However, they often depend on a fixed scanning order~\cite{vmamba,mambasurvey}, and rigid paths struggle to align with the complex spatial structures of irregularly shaped lesions. We therefore use state transitions to model the spatial organization of lesion regions without a fixed scan order.

To address this, in this paper we propose a non-causal, deformation-aware network designed for multi-lesion segmentation in diabetic retinopathy (D$^2$Mamba). Unlike existing DR multi-lesion segmentation methods, our approach employs a non-causal dual state-space mechanism (NC-SSD)~\cite{vssd,mamba2}. This allows each position to leverage global context, enabling the model to efficiently capture spatial relationships among lesion regions while overcoming the unidirectional causal constraints imposed by fixed scanning sequences. Content-dependent state modulation enables the model to distinguish lesion-related cues from irrelevant background, thereby facilitating a diverse global receptive field. In addition, our deformation-aware multi-scan mechanism (DMSM) \cite{defmamba} offers spatial adaptability, which helps mitigate interference from large, irrelevant regions while better conforming to the distinct shapes of individual lesions.

\begin{figure*}[t]
\centering
\includegraphics[width=\textwidth]{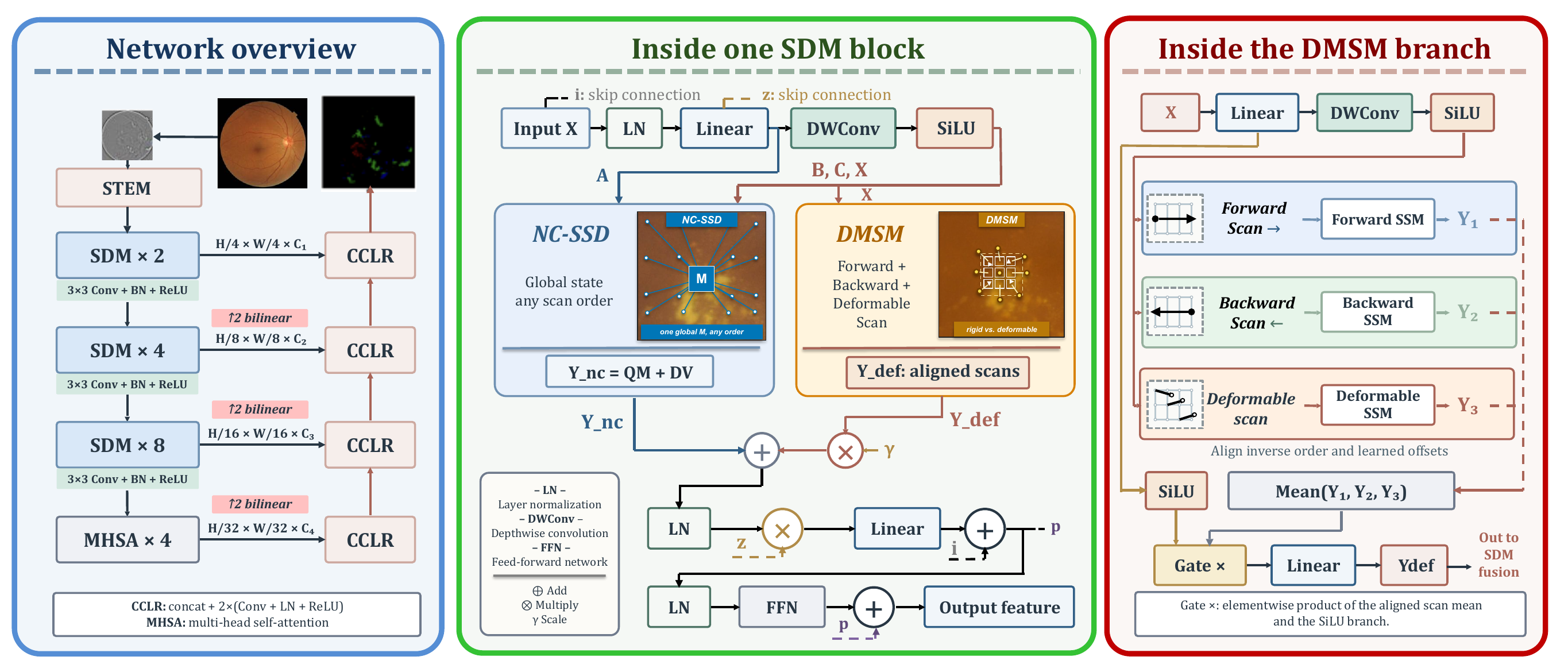}
\caption{\textbf{D$^2$Mamba architecture and SDM Block. }The encoder uses SDM Blocks at three scales; CCLR fuses decoder and skip features. NC-SSD aggregates global context, while DMSM combines forward, backward, and deformable scans.}
\label{fig:network}
\label{fig:block}
\end{figure*}

\section{Method}
\subsection{Network and SDM Block}
Figure~\ref{fig:network} shows a four-stage encoder and progressive CCLR decoder. After a convolutional stem, three SDM stages contain 2, 4, and 8 blocks at widths 64, 128, and 256; the fourth stage contains four multi-head self-attention blocks at width 512. The stages operate at approximately $1/4$, $1/8$, $1/16$, and $1/32$ input resolution. CCLR fuses upsampled decoder and encoder features. A third-stage auxiliary head supervises training; the five-class head produces the inference output.

Within each SDM Block (Fig.~\ref{fig:block}), normalized features of shape $N\times H_s\times W_s\times C_s$ are projected into content and state representations. Flattening the spatial grid gives $L=H_sW_s$ positions. The content projection has width $2C_s$, partitioned into heads of width $d=64$; the state features have width $r=64$. For one sample and head, let $V\in\mathbb R^{L\times d}$ be content and $K,Q\in\mathbb R^{L\times r}$ be state features shared across the value heads. NC-SSD computes a global state summary and projects it back to every position:
\begin{equation}
 M=K^\top\operatorname{diag}(\rho)V,\quad
 Y_{\rm nc}=QM+DV,\quad \rho=-A\Delta,
 \label{eq:ncssd}
\end{equation}
Here $\Delta$ is an input-dependent positive step, and $A$ and $D$ are learned head parameters. The $r\times d$ summary $M$ aggregates all $L$ positions without a causal mask or dense $L\times L$ attention matrix. $Q$ maps $M$ to $L\times d$; concatenating heads gives $Y_{\rm nc}\in\mathbb R^{N\times L\times2C_s}$.

DMSM predicts offsets on the two-dimensional grid and bilinearly samples features. It scans them forward, backward, and in a deformable order set by learned scores. The backward output is reversed, and the deformable output is restored by the inverse learned permutation before the three aligned streams are averaged, gated, and projected. With state width 16, DMSM produces $Y_{\rm def}\in\mathbb R^{N\times L\times2C_s}$, matching $Y_{\rm nc}$ for per-channel fusion:
\begin{equation}
Y=Y_{\rm nc}+\gamma\odot Y_{\rm def},\qquad \gamma_c^{(0)}=10^{-3}.
\label{eq:fusion}
\end{equation}
The learned channel scale $\gamma$ controls the deformable contribution; both branches are trained end to end.

\subsection{Multi-scale Decoding}
Starting from the deepest feature, CCLR aligns the current decoder map with the next encoder skip and concatenates them:
\begin{equation}
D_s=\phi_s\!\left([\mathcal U_s(D_{s+1}),P_s(F_s)]\right).
\label{eq:decoder}
\end{equation}
Here $F_s$ is the encoder feature at stage $s$, $P_s$ adjusts its channels, and $\mathcal U_s$ bilinearly resizes the decoder feature to the size of $F_s$. The operator $\phi_s$ applies two successive groups, each with a $3\times3$ convolution, LayerNorm2d, and ReLU, to the concatenated features.

\begin{figure*}[t]
\centering
\includegraphics[width=0.95\textwidth]{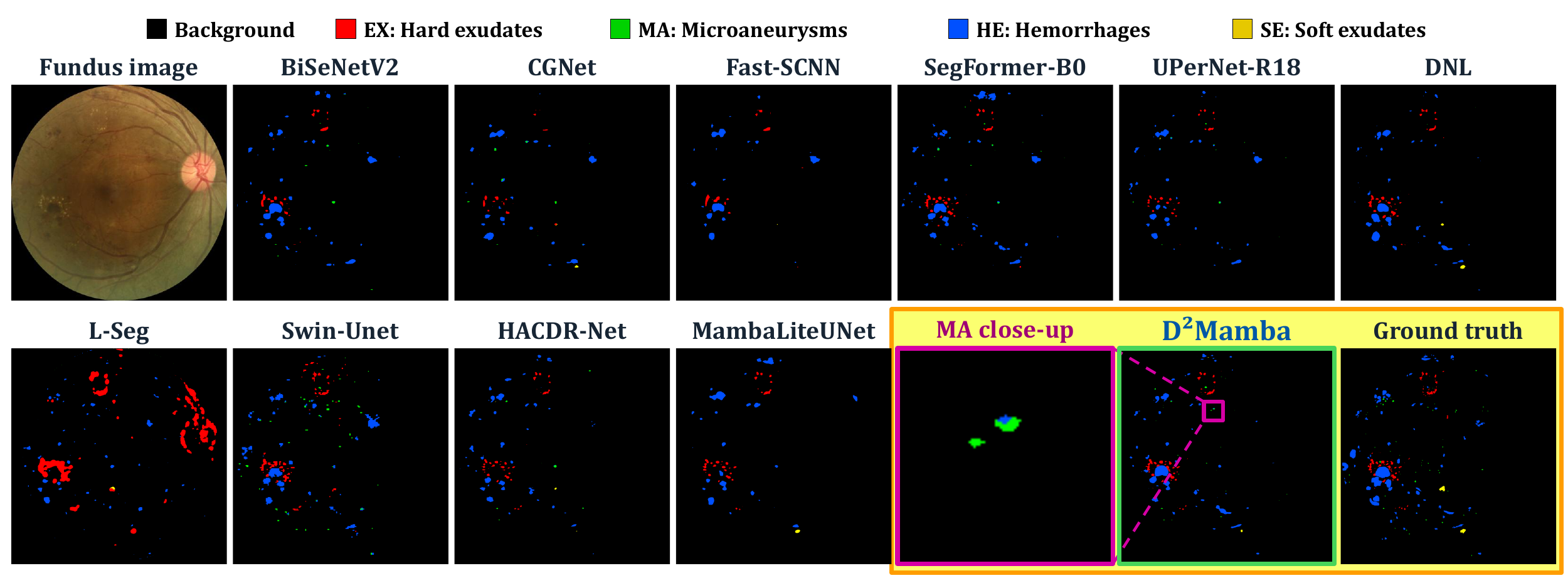}
    \caption{\textbf{Qualitative comparison on DDR image 007-6148-300.} The panels show the fundus image, baseline predictions, D$^2$Mamba, and ground truth. The magenta box marks the region enlarged in the adjacent MA close-up. }
\label{fig:qualitative}
\end{figure*}

\begin{table*}[t]
\centering\footnotesize
\caption{\textbf{DDR five-fold IoU results:} mean $\pm$ population standard deviation (\%). EX, MA, SE, and HE denote hard exudates, microaneurysms, soft exudates, and hemorrhages. The macro mean excludes background. Bold marks the best mean; underlining marks the second best. Rows are grouped by backbone architecture; the Purpose icon gives the target stated in each cited paper: \protect\icFast\ general-purpose real-time/efficient, \protect\icGeneral\ general-purpose scene parsing, \protect\icMedical\ medical image segmentation.}
\label{tab:results}
\setlength{\tabcolsep}{3.6pt}
\renewcommand{\arraystretch}{1.04}
\begin{tabular}{@{}l l c *{5}{c}@{}}
\toprule
\multirow{2}{*}{Model} & \multirow{2}{*}{Key design} & \multirow{2}{*}{Purpose} & \multicolumn{5}{c}{IoU (\%)} \\
\cmidrule(l){4-8}
 & & & Mean & EX & MA & SE & HE \\
\midrule
\multicolumn{8}{@{}l}{\textit{CNN}} \\
\rowcolor{gray!8} L-Seg~\cite{lseg} & multi-scale fusion + bin loss & \icMedical & 11.30 $\pm$ 1.60 & 13.01 $\pm$ 2.71 & 2.57 $\pm$ 3.63 & 18.50 $\pm$ 3.69 & 11.14 $\pm$ 1.55 \\
Fast-SCNN~\cite{fastscnn} & learning-to-downsample, two-branch & \icFast & 20.52 $\pm$ 1.40 & 31.03 $\pm$ 1.43 & 0.73 $\pm$ 0.50 & 27.15 $\pm$ 1.52 & 23.18 $\pm$ 5.74 \\
\rowcolor{gray!8} CGNet~\cite{cgnet} & context-guided block & \icFast & 23.48 $\pm$ 1.98 & 33.47 $\pm$ 2.68 & 4.28 $\pm$ 1.54 & 25.78 $\pm$ 3.46 & 30.39 $\pm$ 6.41 \\
BiSeNetV2~\cite{bisenetv2} & detail + semantic branches & \icFast & 26.35 $\pm$ 1.49 & 34.02 $\pm$ 2.73 & 8.35 $\pm$ 1.32 & 32.23 $\pm$ 3.24 & 30.81 $\pm$ 5.02 \\
\rowcolor{gray!8} UPerNet-R18~\cite{upernet,resnet} & unified perceptual parsing & \icGeneral & 27.85 $\pm$ 1.93 & 36.90 $\pm$ 3.25 & \underline{11.70} $\pm$ 2.62 & 33.13 $\pm$ 2.18 & 29.68 $\pm$ 6.91 \\
DNL~\cite{dnl} & disentangled non-local & \icGeneral & 27.88 $\pm$ 2.74 & 35.49 $\pm$ 4.78 & 0.39 $\pm$ 0.46 & 41.39 $\pm$ 3.41 & \textbf{34.23} $\pm$ 8.30 \\
\rowcolor{gray!8} HACDR-Net~\cite{hacdr} & heterogeneous-aware convolution & \icMedical & 28.97 $\pm$ 1.33 & 39.01 $\pm$ 1.87 & 9.86 $\pm$ 1.34 & 35.44 $\pm$ 3.80 & 31.56 $\pm$ 5.47 \\
\midrule
\multicolumn{8}{@{}l}{\textit{Transformer}} \\
\rowcolor{gray!8} SegFormer-B0~\cite{segformer} & hierarchical Transformer + MLP decoder & \icFast & 28.84 $\pm$ 1.18 & \underline{39.82} $\pm$ 2.47 & 8.88 $\pm$ 1.82 & 33.98 $\pm$ 3.50 & 32.68 $\pm$ 5.37 \\
Swin-Unet~\cite{swinunet} & Swin U-shaped encoder and decoder & \icMedical & \underline{30.88} $\pm$ 1.77 & 39.61 $\pm$ 2.03 & 10.04 $\pm$ 0.66 & \underline{41.53} $\pm$ 3.77 & 32.36 $\pm$ 4.14 \\
\midrule
\multicolumn{8}{@{}l}{\textit{Mamba / SSM}} \\
\rowcolor{gray!8} MambaLiteUNet~\cite{mambaliteunet} & Mamba U-Net + cross-gated attention & \icMedical & 22.94 $\pm$ 2.23 & 33.86 $\pm$ 5.17 & 3.82 $\pm$ 2.45 & 26.31 $\pm$ 13.83 & 27.76 $\pm$ 4.95 \\
\brow \textbf{D$^2$Mamba} & NC-SSD + deformable DMSM & \icMedical & \textbf{33.87 $\pm$ 2.78} & \textbf{41.14 $\pm$ 3.47} & \textbf{18.29 $\pm$ 0.75} & \textbf{43.09 $\pm$ 4.96} & \underline{32.98} $\pm$ 5.76 \\
\bottomrule
\end{tabular}
\vspace{0.7em}

\begin{minipage}[b]{0.58\textwidth}\centering
\includegraphics[width=\linewidth]{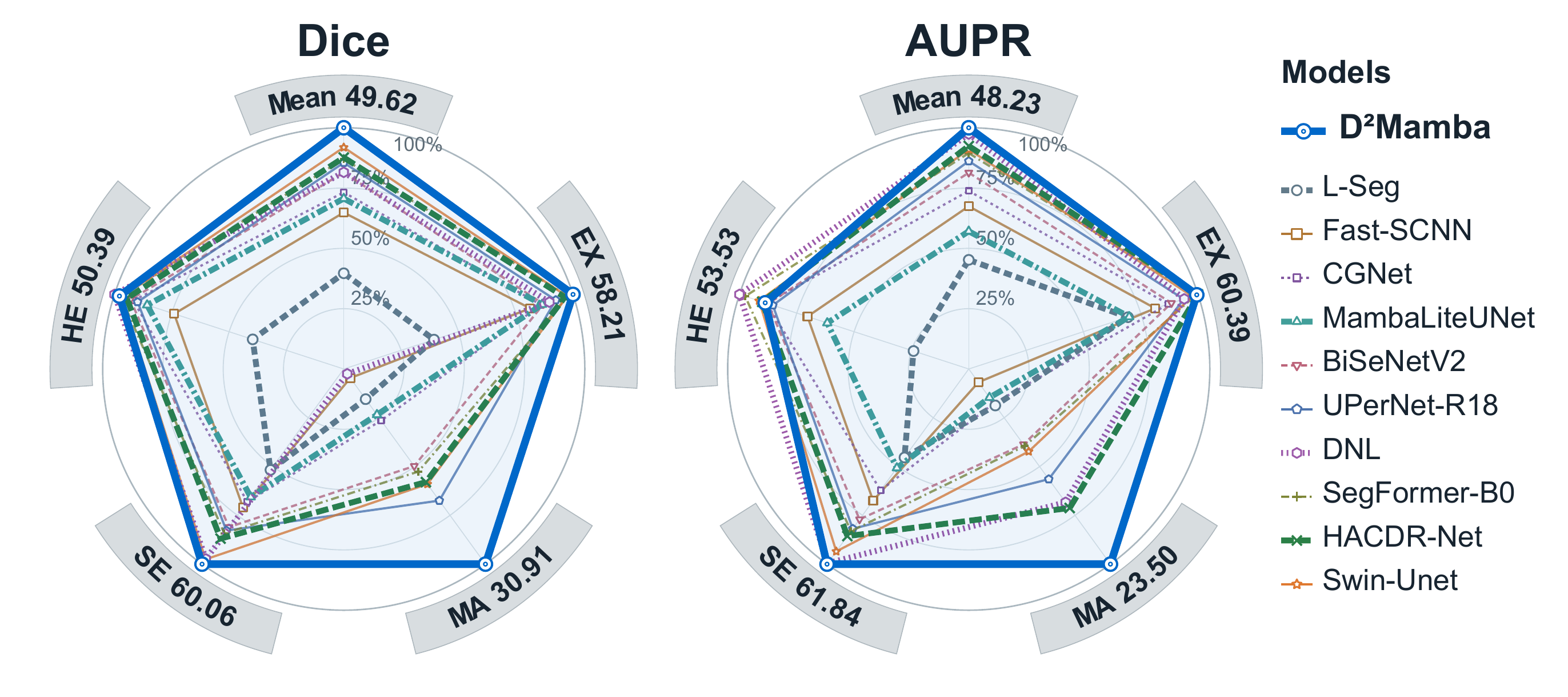}\\[4pt]
{\footnotesize (a) Per-class radar plot}
\end{minipage}\hfill
\begin{minipage}[b]{0.39\textwidth}\centering\footnotesize
\setlength{\tabcolsep}{6pt}\renewcommand{\arraystretch}{1.12}
\begin{tabular}{@{}lccc@{}}
\toprule
Metric (\%) & VSSD & D$^2$Mamba & $\Delta$ \\
\midrule
\multicolumn{4}{@{}l}{\textit{Macro}} \\
\rowcolor{gray!8} mIoU & 36.09 & \textbf{37.36} & \cellcolor{green!12}+1.27 \\
mDice & 51.95 & \textbf{53.37} & \cellcolor{green!12}+1.43 \\
\rowcolor{gray!8} mAUPR & 51.74 & \textbf{52.07} & \cellcolor{green!12}+0.33 \\
\midrule
\multicolumn{4}{@{}l}{\textit{Class-wise IoU}} \\
\rowcolor{gray!8} EX & 44.61 & \textbf{46.78} & \cellcolor{green!12}+2.16 \\
MA & 16.99 & \textbf{19.60} & \cellcolor{green!12}+2.60 \\
\rowcolor{gray!8} SE & 42.76 & \textbf{46.88} & \cellcolor{green!12}+4.12 \\
HE & \textbf{39.99} & 36.20 & \cellcolor{red!12}$-$3.79 \\
\bottomrule
\end{tabular}\\[4pt]
{\footnotesize (b) Ablation study}
\end{minipage}
\captionof{figure}{\textbf{(a)} DDR five-fold mean Dice and AUPR for the four lesion classes and their macro mean. Each grey outer band gives the highest score (\%) on that axis; the concentric rings show 25\% to 100\% of that score. Lines show fold means without error bars; macro means exclude background. \textbf{(b)} DDR test-set ablation of D$^2$Mamba against the NC-SSD-only VSSD variant; $\Delta$ is D$^2$Mamba minus VSSD before rounding, and bold marks the better value. Table~\ref{tab:results} reports five-fold means.}
\label{fig:radar}
\end{table*}

\section{Experiments and Results}
\subsection{Protocol}
We use 757 lesion-annotated DDR images~\cite{ddr} in reconstructed image-level five-fold splits. Each fold contains 453 to 455 training images, 151 to 152 validation images, and 151 to 152 test images. Validation mDice selects the checkpoint; test scores are then averaged equally across folds. We report population standard deviations and exclude background from four-lesion macro scores. IoU and Dice are computed from class-wise confusion counts pooled within each test fold, while AUPR is computed from continuous softmax scores using 1000 histogram bins. We observed that the official split may allow existing models to rely on shortcut learning, so we employed a five-fold cross-validation. Our results cannot be directly compared with those reported in other studies.

We train with PyTorch and MMSegmentation on one NVIDIA RTX 4090D without pretrained initialization. Preprocessing removes black borders and combines each image with a Gaussian-blurred copy for fundus background correction, followed by RGB conversion and ImageNet normalization. Images and labels are resized to $1028\times1028$ using bilinear and nearest-neighbor interpolation, respectively. During training, images and labels share random scaling by factors from $0.5$ to $2.0$, cropping, and flipping; photometric distortion is applied only to images. Validation and testing use deterministic preprocessing. AdamW uses batch size 4, an initial learning rate of $6\times10^{-5}$, and weight decay of 0.01. Each fold is trained for 100000 optimizer updates, with 1500 warm-up updates followed by polynomial learning-rate decay. Validation and checkpointing occur every 2000 updates. The training budget is measured in optimizer updates rather than epochs. All baselines use the same folds.

\subsection{Quantitative Comparison}
Table~\ref{tab:results} compares D$^2$Mamba with ten re-implemented baselines on the same DDR folds. Its mIoU of 33.87\% exceeds Swin-Unet by 2.99 points and MambaLiteUNet by 10.93 points, indicating that modeling the spatial organization of lesions matters more than adopting Mamba alone. Fig.~\ref{fig:radar}(a) shows class-wise Dice and AUPR; D$^2$Mamba also has the highest macro means.

D$^2$Mamba reaches 49.62\% mDice versus 45.57\% for Swin-Unet, and 48.23\% mAUPR versus 46.98\% for DNL, gains of 4.05 and 1.25 points.

\subsection{Class-wise Behavior}

The largest gain is for MA: D$^2$Mamba reaches 18.29\% IoU and 30.91\% Dice, against UPerNet-R18's 11.70\% and 20.85\%, and 23.50\% AUPR against HACDR-Net's 16.74\%. MAs span only a few pixels and are the class most easily suppressed by downsampling and large-scale aggregation, so this gain is the most direct evidence for the intended behavior of the SDM block. D$^2$Mamba also has the highest IoU and Dice for EX and SE, whereas DNL leads on HE in IoU (34.23\% vs. 32.98\%) and Dice (50.39\% vs. 49.30\%). Figure~\ref{fig:qualitative} shows one test case, including the MA close-up.

\subsection{Ablation}

Ablations are conducted on DDR test set under the same training protocol (Fig.~\ref{fig:radar}(b)). Adding DMSM to the NC-SSD-only variant raises mIoU from 36.09\% to 37.36\%, mDice from 51.95\% to 53.37\%, and mAUPR from 51.74\% to 52.07\%. The consistent gains on EX, MA, and SE indicate that deformable scanning complements the global NC-SSD state with spatially adaptive sampling of small and irregular lesions. HE IoU, however, drops by 3.79 points, consistent with HE being the only class where D$^2$Mamba trails a baseline in Table~\ref{tab:results}. A fixed fusion scale appears less suited to large diffuse hemorrhages, and adapting the deformable contribution to lesion scale is left for future work.

\section{Conclusion}

This paper presented D$^2$Mamba for multi-lesion segmentation in diabetic retinopathy. Its SDM block models the spatial organization of lesion regions by coupling non-causal state space duality, which aggregates global context without a fixed scan order, with deformable Mamba scanning, which adapts sampling to lesion shape. In five-fold evaluation on DDR under an identical training protocol, D$^2$Mamba achieved the highest macro IoU, Dice, and AUPR among the compared methods, with the largest gain on microaneurysms, the smallest and most easily missed lesion class. Future work will adapt the deformable contribution to lesion scale and validate the method on additional fundus datasets.

\newpage

\section{Compliance with Ethical Standards}

This research study was conducted retrospectively using human subject data made available in open access by DDR fundus images and annotations dataset~\cite{ddr}. Ethical approval was not required as confirmed by the license attached with the open access data.

\section{Acknowledgments}
No funding was received for this work. The authors declare no relevant financial or non-financial conflicts of interest.

\begingroup
\footnotesize
\bibliographystyle{IEEEbib}
\bibliography{d2mamba_refs}

@article{mamba,
 author={Albert Gu and Tri Dao},
 title={{Mamba}: Linear-Time Sequence Modeling with Selective State Spaces},
 journal={arXiv preprint arXiv:2312.00752}, year={2023},
 url={https://arxiv.org/abs/2312.00752}
}

@article{lseg,
 author={Song Guo and Tao Li and Hong Kang and Ning Li and Yujun Zhang and Kai Wang},
 title={{L-Seg}: An End-to-End Unified Framework for Multi-Lesion Segmentation of Fundus Images},
 journal={Neurocomputing}, volume={349}, pages={52--63}, year={2019},
 doi={10.1016/j.neucom.2019.04.019}
}

@article{fastscnn,
 author={Rudra P. K. Poudel and Stephan Liwicki and Roberto Cipolla},
 title={{Fast-SCNN}: Fast Semantic Segmentation Network},
 journal={arXiv preprint arXiv:1902.04502}, year={2019},
 url={https://arxiv.org/abs/1902.04502}
}

@article{cgnet,
 author={Tianyi Wu and Sheng Tang and Rui Zhang and Juan Cao and Yongdong Zhang},
 title={{CGNet}: A Light-Weight Context Guided Network for Semantic Segmentation},
 journal={IEEE Transactions on Image Processing}, volume={30}, pages={1169--1179}, year={2021},
 doi={10.1109/TIP.2020.3042065}
}

@article{bisenetv2,
 author={Changqian Yu and Changxin Gao and Jingbo Wang and Gang Yu and Chunhua Shen and Nong Sang},
 title={{BiSeNet V2}: Bilateral Network with Guided Aggregation for Real-Time Semantic Segmentation},
 journal={International Journal of Computer Vision}, volume={129}, pages={3051--3068}, year={2021},
 doi={10.1007/s11263-021-01515-2}
}

@inproceedings{upernet,
 author={Tete Xiao and Yingcheng Liu and Bolei Zhou and Yuning Jiang and Jian Sun},
 title={Unified Perceptual Parsing for Scene Understanding},
 booktitle={European Conference on Computer Vision (ECCV)}, pages={432--448}, year={2018},
 doi={10.1007/978-3-030-01228-1_26}
}

@inproceedings{resnet,
 author={Kaiming He and Xiangyu Zhang and Shaoqing Ren and Jian Sun},
 title={Deep Residual Learning for Image Recognition},
 booktitle={IEEE Conference on Computer Vision and Pattern Recognition (CVPR)},
 pages={770--778}, year={2016}, doi={10.1109/CVPR.2016.90}
}

@inproceedings{dnl,
 author={Minghao Yin and Zhuliang Yao and Yue Cao and Xiu Li and Zheng Zhang and Stephen Lin and Han Hu},
 title={Disentangled Non-Local Neural Networks},
 booktitle={European Conference on Computer Vision (ECCV)}, pages={191--207}, year={2020},
 doi={10.1007/978-3-030-58555-6_12}
}

@inproceedings{swinunet,
 author={Hu Cao and Yueyue Wang and Joy Chen and Dongsheng Jiang and Xiaopeng Zhang and Qi Tian and Manning Wang},
 title={{Swin-Unet}: Unet-Like Pure Transformer for Medical Image Segmentation},
 booktitle={Computer Vision -- ECCV 2022 Workshops}, pages={205--218}, year={2023},
 doi={10.1007/978-3-031-25066-8_9}
}

@article{mambaliteunet,
 author={Md Maklachur Rahman and Soon Ki Jung and Tracy Hammond},
 title={{MambaLiteUNet}: Cross-Gated Adaptive Feature Fusion for Robust Skin Lesion Segmentation},
 journal={arXiv preprint arXiv:2604.20286}, year={2026},
 url={https://arxiv.org/abs/2604.20286}
}

@article{ddr,
 author={Tao Li and Yingqi Gao and Kai Wang and Song Guo and Hanruo Liu and Hong Kang},
 title={Diagnostic Assessment of Deep Learning Algorithms for Diabetic Retinopathy Screening},
 journal={Information Sciences}, volume={501}, pages={511--522}, year={2019}, doi={10.1016/j.ins.2019.06.011}
}

@inproceedings{segformer,
 author={Enze Xie and Wenhai Wang and Zhiding Yu and Anima Anandkumar and Jose M. Alvarez and Ping Luo},
 title={{SegFormer}: Simple and Efficient Design for Semantic Segmentation with Transformers},
 booktitle={Advances in Neural Information Processing Systems}, volume={34}, pages={12077--12090}, year={2021}
}

@article{vssd,
 author={Yuheng Shi and Minjing Dong and Mingjia Li and Chang Xu},
 title={{VSSD}: Vision Mamba with Non-Causal State Space Duality},
 journal={arXiv preprint arXiv:2407.18559}, year={2024}, url={https://arxiv.org/abs/2407.18559}
}

@article{defmamba,
 author={Leiye Liu and Miao Zhang and Jihao Yin and Tingwei Liu and Wei Ji and Yongri Piao and Huchuan Lu},
 title={{DefMamba}: Deformable Visual State Space Model},
 journal={arXiv preprint arXiv:2504.05794}, year={2025}, url={https://arxiv.org/abs/2504.05794}
}

@inproceedings{hacdr,
 author={QiHao Xu and Xiaoling Luo and Chao Huang and Chengliang Liu and Jie Wen and Jialei Wang and Yong Xu},
 title={{HACDR-Net}: Heterogeneous-Aware Convolutional Network for Diabetic Retinopathy Multi-Lesion Segmentation},
 booktitle={Proceedings of the AAAI Conference on Artificial Intelligence}, volume={38}, pages={6342--6350}, year={2024}, doi={10.1609/aaai.v38i6.28453}
}

@article{nnunet,
 author={Fabian Isensee and Paul F. Jaeger and Simon A. A. Kohl and Jens Petersen and Klaus H. Maier-Hein},
 title={{nnU-Net}: A Self-Configuring Method for Deep Learning-Based Biomedical Image Segmentation},
 journal={Nature Methods}, volume={18}, pages={203--211}, year={2021},
 doi={10.1038/s41592-020-01008-z}
}

@inproceedings{unet,
 author={Olaf Ronneberger and Philipp Fischer and Thomas Brox},
 title={{U-Net}: Convolutional Networks for Biomedical Image Segmentation},
 booktitle={Medical Image Computing and Computer-Assisted Intervention (MICCAI)}, pages={234--241}, year={2015},
 doi={10.1007/978-3-319-24574-4_28}
}

@article{vmamba,
 author={Yue Liu and Yunjie Tian and Yuzhong Zhao and Hongtian Yu and Lingxi Xie and Yaowei Wang and Qixiang Ye and Jianbin Jiao and Yunfan Liu},
 title={{VMamba}: Visual State Space Model},
 journal={arXiv preprint arXiv:2401.10166}, year={2024}
}

@article{mamba2,
 author={Tri Dao and Albert Gu},
 title={Transformers are {SSMs}: Generalized Models and Efficient Algorithms Through Structured State Space Duality},
 journal={arXiv preprint arXiv:2405.21060}, year={2024}
}

@article{teo2021,
 author={Zhen Ling Teo and Yih-Chung Tham and Marco Yu and Miao Li Chee and Tyler Hyungtaek Rim and Ning Cheung and others},
 title={Global Prevalence of Diabetic Retinopathy and Projection of Burden through 2045},
 journal={Ophthalmology},
 volume={128},
 number={11},
 pages={1580--1591},
 year={2021},
 doi={10.1016/j.ophtha.2021.04.027}
}

@article{abramoff2018,
 author={Michael D. Abr{\`a}moff and Philip T. Lavin and Michele Birch and Nilay Shah and James C. Folk},
 title={Pivotal trial of an autonomous {AI}-based diagnostic system for detection of diabetic retinopathy in primary care offices},
 journal={npj Digital Medicine},
 volume={1},
 number={1},
 year={2018},
 doi={10.1038/s41746-018-0040-6}
}

@article{gulshan2016,
 author={Varun Gulshan and Lily Peng and Marc Coram and Martin C. Stumpe and Derek Wu and Arunachalam Narayanaswamy and others},
 title={Development and Validation of a Deep Learning Algorithm for Detection of Diabetic Retinopathy in Retinal Fundus Photographs},
 journal={JAMA},
 volume={316},
 number={22},
 pages={2402},
 year={2016},
 doi={10.1001/jama.2016.17216}
}

@article{ting2017,
 author={Daniel Shu Wei Ting and Carol Yim-Lui Cheung and Gilbert Lim and Gavin Siew Wei Tan and Nguyen D. Quang and Alfred Gan and others},
 title={Development and Validation of a Deep Learning System for Diabetic Retinopathy and Related Eye Diseases Using Retinal Images From Multiethnic Populations With Diabetes},
 journal={JAMA},
 volume={318},
 number={22},
 pages={2211},
 year={2017},
 doi={10.1001/jama.2017.18152}
}

@article{idrid2020,
 author={Prasanna Porwal and Samiksha Pachade and Manesh Kokare and Girish Deshmukh and Jaemin Son and Woong Bae and others},
 title={{IDRiD}: Diabetic Retinopathy -- Segmentation and Grading Challenge},
 journal={Medical Image Analysis},
 volume={59},
 pages={101561},
 year={2020},
 doi={10.1016/j.media.2019.101561}
}

@article{idrid2018,
 author={Prasanna Porwal and Samiksha Pachade and Ravi Kamble and Manesh Kokare and Girish Deshmukh and Vivek Sahasrabuddhe and others},
 title={{I}ndian {D}iabetic Retinopathy Image Dataset ({IDRiD}): A Database for Diabetic Retinopathy Screening Research},
 journal={Data},
 volume={3},
 number={3},
 pages={25},
 year={2018},
 doi={10.3390/data3030025}
}

@article{retfound,
 author={Yukun Zhou and Mark A. Chia and Siegfried K. Wagner and Murat S. Ayhan and Dominic J. Williamson and Robbert R. Struyven and others},
 title={A foundation model for generalizable disease detection from retinal images},
 journal={Nature},
 volume={622},
 number={7981},
 pages={156--163},
 year={2023},
 doi={10.1038/s41586-023-06555-x}
}

@article{cheung2010,
 author={Ning Cheung and Paul Mitchell and Tien Yin Wong},
 title={{D}iabetic retinopathy},
 journal={The Lancet},
 volume={376},
 number={9735},
 pages={124--136},
 year={2010},
 doi={10.1016/s0140-6736(09)62124-3}
}

@article{litjens2017,
 author={Geert Litjens and Thijs Kooi and Babak Ehteshami Bejnordi and Arnaud Arindra Adiyoso Setio and Francesco Ciompi and Mohsen Ghafoorian and others},
 title={A survey on deep learning in medical image analysis},
 journal={Medical Image Analysis},
 volume={42},
 pages={60--88},
 year={2017},
 doi={10.1016/j.media.2017.07.005}
}

@article{mambasurvey,
 author={Rui Xu and Shu Yang and Yihui Wang and Yu Cai and Bo Du and Hao Chen},
 title={Visual {Mamba}: A Survey and New Outlooks},
 journal={arXiv preprint arXiv:2404.18861}, year={2024}
}

@article{umamba,
  title   = {U-Mamba: Enhancing Long-range Dependency for Biomedical Image Segmentation},
  author  = {Ma, Jun and Li, Feifei and Wang, Bo},
  journal = {arXiv preprint arXiv:2401.04722},
  year    = {2024}
}
\endgroup
\end{document}